%% file: sigspatial2019.tex
\documentclass[sigconf]{acmart}

\copyrightyear{2019}
\acmYear{2019}
\acmConference[SIGSPATIAL '19]{27th ACM SIGSPATIAL International Conference on Advances in Geographic Information Systems}{November 5--8, 2019}{Chicago, IL, USA}
\acmBooktitle{27th ACM SIGSPATIAL International Conference on Advances in Geographic Information Systems (SIGSPATIAL '19), November 5--8, 2019, Chicago, IL, USA}
\acmPrice{15.00}
\acmDOI{10.1145/3347146.3359381}
\acmISBN{978-1-4503-6909-1/19/11}

\usepackage[ruled]{algorithm2e}
\usepackage[tight,footnotesize]{subfigure}
\usepackage{hhline}

\newcommand{\brm}[1]{\boldsymbol{\mathrm{#1}}} 
\newcommand{\ls}[1]{{\Vert#1\Vert}_{2}^{2}} 
\newcommand{\lfrob}[1]{{\Vert#1\Vert}_{F}} 
\newcommand{\lfrobsq}[1]{{\Vert#1\Vert}_{F}^{2}} 
\newcommand{\lnorm}[2]{{\Vert#1\Vert}_{#2}} 
\newcommand{\lone}[1]{\lnorm{#1}{1}} 
\newcommand{\ltwo}[1]{\lnorm{#1}{2}} 
\newcommand{\argmin}[1]{\underset{#1}{\operatorname{argmin}}\;}

\newcommand{\prox}{\operatorname{prox}}
\DeclareMathAlphabet{\mathdata}{OMS}{cmsy}{m}{n}

\def\X{\brm{X}}

\def\Y{\brm{Y}}
\def\v{\brm{v}}

\def\w{\brm{w}}
\def\W{\brm{W}}

\def\I{\brm{I}}

\def\U{\brm{U}}

\def\q{\brm{q}}
\def\Q{\brm{Q}}

\begin{document}

\title{\textit{TITAN}: A Spatiotemporal Feature Learning Framework for Traffic Incident Duration Prediction}

\author{Kaiqun Fu$^1$, Taoran Ji$^1$, Liang Zhao$^2$, Chang-Tien Lu$^1$}
\email{{fukaiqun, jtr}@vt.edu, lzhao9@gmu.edu, ctlu@vt.edu}
\affiliation{%
  \institution{$^1$Virginia Tech}
}
\affiliation{%
  \institution{$^2$George Mason University}
}

\renewcommand{\shortauthors}{Fu, et al.}

\begin{abstract}

Critical incident stages identification and reasonable prediction of traffic
incident duration are essential in traffic incident management. In this paper,
we propose a traffic incident duration prediction model that simultaneously
predicts the impact of the traffic incidents and identifies the critical
groups of temporal features via a multi-task learning framework. First, we
formulate a sparsity optimization problem that extracts low-level temporal
features based on traffic speed readings and then generalizes higher level
features as phases of traffic incidents. Second, we propose novel constraints
on feature similarity exploiting prior knowledge about the spatial
connectivity of the road network to predict the incident duration. The
proposed problem is challenging to solve due to the orthogonality constraints,
non-convexity objective, and non-smoothness penalties. We develop an algorithm
based on the alternating direction method of multipliers (ADMM) framework to
solve the proposed formulation. Extensive experiments and comparisons to other
models on real-world traffic data and traffic incident records justify the
efficacy of our model.

\end{abstract}




\maketitle

\input{sections/Introduction.tex}
\input{sections/Related.tex}
\input{sections/ProbState.tex}
\input{sections/Models.tex}
\input{sections/Algorithm.tex}
\input{sections/Experiment.tex}
\input{sections/Conclusion.tex}

\bibliographystyle{ACM-Reference-Format}
\bibliography{references}

\end{document}

%% file: sections/Introduction.tex
\section{Introduction}
\label{sec:introduction}

The studies of early detecting the traffic incidents and estimating the impact
of the non-recurrent congestions caused by traffic incidents have become
increasingly important research topics due to the significant social and
economic losses generated. A one-minute reduction on congestion duration
produces a 65 US dollars gain per incident~\cite{adler2013road}. Although
non-recurrent congestion is hard to predict due to its nature of randomness,
the studies on impact and duration of the traffic incidents are still ones of
the major focuses for the traffic operators. The vast deployment of
transportation traffic speed sensors and Traffic Incident Management Systems
(TIMS) make the traffic speed data and traffic incident records ubiquitously
accessible for the transportation operators. With the abundance of the traffic
data sources, an efficient multi-task learning model can be implemented to
provide an accurate prediction on incident duration.

\begin{figure*}[htpb!]
\centering
\includegraphics[width=\linewidth]{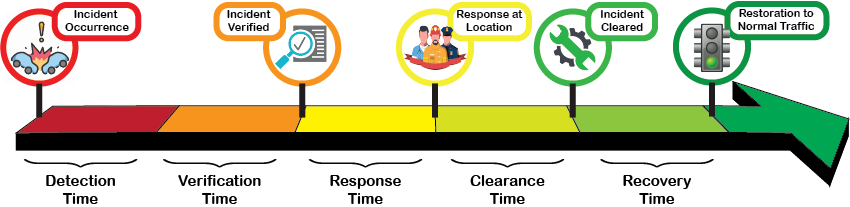} 
\caption{From the perspectives of traffic management and transportation operations, the life cycle of a traffic incident is separated into five stages: Detection, Verification, Response, Clearance, and Recovery}
\label{fig:intro_2}
\end{figure*}

Incident duration is the time elapsed from the incident occurrence until all
evidence of the incident has been removed from the incident scene. From the
perspective of traffic management and operation, the life cycle of a traffic
incident is split into five stages: Detection, Verification, Response,
Clearance, and Recovery~\cite{ozbay1999incident}. Figure~\ref{fig:intro_2}
shows the life cycle of a traffic incident. However, the five-stage life cycle
separation cannot be used directly as the temporal features for the traffic
incident duration prediction. To accurately estimate the duration of a traffic
incident in its early stages, the transportation operators and first
responders encounter three major challenges: \textbf{1) No explicit high-level
temporal features}: Although the conventional five-stage life cycle separation
is effective for the purposes of traffic management, such five-stages cannot
be considered as temporal features in traffic incident duration prediction
task. It is important to group the critical time point features in the early
stages of the incident forming higher level time periods that can perform as a
better indicator for predicting the incident duration. \textbf{2) Hard to
predict the influence of incident}: In the research field of Traffic Incident
Management, one of the most essential tasks is to estimate the impact of the
traffic incident in terms of its temporal duration at early stages. However,
the performances of the conventional time series based methods are limited by
their incapability of identifying higher level temporal features. \textbf{3)
Spatial connectivity of the road networks is rarely considered}: The traffic
congestion cascades within the road network. As a consequence, the traffic
patterns of incidents in their early stages are similar when the traffic
incidents are topologically closer from the perspective of the road networks.
Traffic incidents that are spatiotemporally closer should share more similar
traffic speed patterns. However, this spatial correlation between traffic
incidents is rarely considered in the previous studies~\cite{li2018overview}. 


The existing methods are mostly infeasible to solve these challenges. Current
feature learning methods such as $\ell_1$-norm regularized methods such as
Lasso~\cite{tibshirani2005sparsity} have properties in terms of feature
selection. However, strong assumptions on the design matrix are
required~\cite{zhou2013modeling}. Zhan et al.~\cite{zhan2011prediction}
propose an M5P tree algorithm to predict the clearance time of traffic
incident based on the geometric, and traffic features. Feature learning
algorithms for biomarker identification~\cite{zhou2013feafiner} and social
event indicators~\cite{zhao2018distant} are proved to be effective while
finding higher level features. However, most of them focus on learning
important feature sets from attributes and does not apply to our encountered
problem due to expensive computation. In these studies, they considered the
duration of an incident to quantify the impact. However, their quantification
strategies are designed to capture the one-time impact of the incident,
instead of the time-varying nature of impact at different locations.
Multi-task learning based spatiotemporal model plays an important role while
considering the connectivity of the road networks.  Multi-task based
spatiotemporal models focus on regression and classification problems such as
county income prediction~\cite{zhang2017spatiotemporal}, social unrest event
forecasting~\cite{zhao2016hierarchical}, and even service disruption detection
for transit networks~\cite{ji2018multi}. However, none of the previously
proposed methods is capable of modeling the spatial connectivity between
features at a higher level. Therefore, most of the existing models are not
suitable for our traffic incident duration prediction problem. 

To address these challenges, we propose a \textbf{T}raffic
\textbf{I}nciden\textbf{T} Dur\textbf{A}tion Predictio\textbf{N}
(\textit{TITAN}) model based on both sparse feature learning and multi-task
learning framework. Our main contributions are:

$\bullet$ \textbf{Formulating a novel machine learning framework for traffic incident duration prediction using temporal features}. 
In contrast to existing works, we formulate the problem of traffic
incident duration prediction for transportation systems as a multi-task
supervised learning problem. In the proposed methods, models for different
road segments are learned simultaneously by restricting all road segments
to exploit a common set of features.

$\bullet$ \textbf{Modeling traffic speed similarity among road segments via spatial connectivity in feature space}. 
Based on the cascading nature of the traffic congestion in road networks,
specifically designed constraints are proposed to model traffic speed
similarities among data for spatiotemporally correlated road segments.
These similarities in feature space are driven by spatial connectivity.

$\bullet$ \textbf{Proposing a sparse feature learning process to identify groups of temporal features at a higher level}. 
According to the nature of the traffic incidents, the traffic speed
fluctuation in the early stages of the incidents is always important while
estimating the impact and duration of the traffic incident. In the
proposed model, constraints with sparsity and orthogonality are introduced
to extract grouped important temporal features at a higher level.

$\bullet$ \textbf{Developing an efficient algorithm to train the proposed
model}. The underlying optimization problem of the proposed multi-task model
is a non-smooth, multi-convex, and inequality-constrained problem, which is
challenging to solve. By introducing auxiliary variables, we develop an
effective ADMM- based algorithm to decouple the main problem into several
sub-problems which can be solved by block coordinate descent and proximal
operators.

The rest of our paper is structured as follows. Related works are reviewed in
Section~\ref{sec:related}. In Section~\ref{sec:probstate}, we describe the
problem setup of our work. In Sections~\ref{sec:models}
and~\ref{sec:algorithm}, we present a detailed discussion of our proposed
\textit{TITAN} model for predicting durations of traffic incidents, and its
solution for parameter learning. In Section~\ref{sec:experiment}, extensive
experiment evaluations and comparisons are presented. In the last section, we
discuss our conclusion and directions for future work. 

%% file: sections/Related.tex
\section{Related Works}
\label{sec:related}

In this section, we provide a detailed review of the current state of research
for traffic incident analysis problem. There are several threads of related
work of this paper: traffic incident impacts analysis, urban event
forecasting, and spatiotemporal multi-task learning. 

\textbf{Traffic Impacts Analysis}. The applications of conventional
statistical methods have addressed its effectiveness in the traffic incident
duration time prediction problems. The statistical methods fall into several
branches: Bayesian classifier~\cite{boyles2007naive}, discrete choice
model~\cite{lin2004integration}, linear/non-parametric
regression~\cite{peeta2000providing}, hazard-based duration
model~\cite{nam2000exploratory}. In the recent decade, the Traffic Incident
Management Systems (TIMS) have been deployed by traffic control centers in
various cities and highways to alleviate the influence of traffic incidents on
traffic conditions~\cite{owens2010traffic}. The historical traffic data
obtained corresponds to traffic incidents play an important role in predicting
the traffic incident durations. A new research field based on data-driven
algorithms and supported by real-world traffic data availability has recently
emerged for traffic incident duration prediction with increasing research
popularity. Various data mining and machine learning approaches have been
employed to estimate and predict traffic incident duration time. Some of these
approaches are the following: Lee et al.~\cite{lee2010computerized} proposed a
genetic algorithm on traffic incident duration time prediction problems; Kim
et al. and Zhan et al.~\cite{zhan2011prediction} applied decision trees and
classification tree models on the same problem and achieved improvements;
Valenti et al.~\cite{valenti2010comparative} proposed a support vector machine
related method that utilizes the temporal features of the traffic data;
artificial neural networks~\cite{vlahogianni2013fuzzy} is another highlighted
direction for traffic incident duration prediction. In recent years, the
research field of Intelligent Transportation Systems (ITS) have addressed its
attention towards the hybrid methods~\cite{kim2012development} to predict
traffic incident durations. 

\textbf{Urban Event Forecasting}. To predict and detect the occurrence and
impact the traffic incidents as urban events have received increasing
attention in recent years. A large body of traditional work for event
forecasting has focused on the early detection of events such as
earthquakes~\cite{sakaki2010earthquake}, disease
outbreaks~\cite{zhao2015simnest}, and transit service
disruption~\cite{ji2018multi}, while event forecasting methods predict the
incidence of such events in the future. Temporal events are the major focuses
of the most existing event forecasting methods, with no interest in the
geographical dimension, such as stock market
movements~\cite{bollen2011twitter} and elections~\cite{o2010tweets}. A handful
of works started to address the urban event prediction problem on a
spatiotemporal resolution. For example, Zhao et al.~\cite{zhao2015multi}
proposed a multi-task learning framework that models forecasting tasks in
related geo-locations concurrently and; Gerber et
al.~\cite{gerber2014predicting} utilized a logistic regression model for
spatiotemporal event forecasting, the urban event predictions with true
spatiotemporal resolution. One limitation of these existing studies is that
the temporal dimension is considered to be independent of the spatial
dimension, and any interactions between the two are ignored. Our proposed
\textit{TITAN} model addresses the importance of the topology dimension, which
is derived from the spatial dimension. We propose a multi-task learning
framework with orthogonal constraints to model the interactions between the
temporal and topological dimensions. 

\textbf{Spatiotemporal Multi-task Learning}. Multi-task learning (MTL) refers
to models that learn multiple related tasks simultaneously to improve overall
performance. Recent decades have witnessed proposals for many MTL
approaches~\cite{zhou2011malsar}. Evgeniou et
al.~\cite{evgeniou2004regularized} proposed a regularized MTL formulation that
constrains the models of each task to be close to each other. Task relatedness
can also be modeled by constraining multiple tasks to share a common
underlying structure (e.g., a common set of features)~\cite{argyriou2007multi},
or a common subspace~\cite{ando2005framework}. Zhao et
al.~\cite{zhao2015multi} designed a multi-task learning framework that models
forecasting tasks in related geolocations. MTL approaches have been applied in
many domains including computer vision and biomedical informatics. Our work,
to the best of our knowledge, is the first paper to address the feasibility of
combining multi-task learning and orthogonal regularization techniques to
resolve traffic incident duration prediction and critical phases learning
problems. 

%% file: sections/ProbState.tex
\section{Problem Statement} 
\label{sec:probstate}


Assume that we are given a collection of traffic incidents $\mathdata{I}$ from
the traffic incident management system (TIM). For each traffic incident $i$ in
$\mathdata{I}$, we find the spatially correlated traffic sensor $s$, and its
traffic speed reading at time interval $\tau$: $\v_{s}(\tau)$, the granularity
of the time interval is 1 minute. Given an incident record, and the traffic
speed readings of its corresponding traffic speed sensor, the main objective
of this paper is to predict the future impact of this given incident in terms
of the temporal duration of this traffic incident.

\textbf{Definition I}\label{def:def1}: 
\textit{Traffic speed in detection time and early verification time}. 
Suppose the verification time of the traffic incident is in time interval
$\tau_{v}$, we define and extract two important time periods \textbf{respond
time} (\textit{time between incident occurrence $\tau_{o}$ and incident
verification time $\tau_{v}$}) and \textbf{early verification time} (\textit{a
short period after the traffic incident verification time $\tau_{v}$}) for
feature construction. The traffic
speeds for both time periods are extracted as: \textbf{(1) Traffic speed in
detection time}: the previous $h$ readings: $\v_{s}(\tau_{v}-1),
\v_{s}(\tau_{v}-2), ..., \v_{s}(\tau_{v}-h)$ and \textbf{(2) Traffic speed in
early verification time}: the succeeding $t$ readings $\v_{s}(\tau_{v}),
\v_{s}(\tau_{v}+1), ..., \v_{s}(\tau_{v}+t)$.

Given the collection of traffic incidents, we first filter the collection with
a selection $\Phi$ of arterial roads. This produces the targeted traffic
incidents collection $\mathdata{I}^{+}$. Then based on which traffic incident
takes place at the arterial road, $\mathdata{I}^{+}$ is grouped into
$\{\mathdata{I}^{+}_{r}\}^{r\in \Phi}$, for example,
$\Phi=\{\mathrm{I\mbox{-}270}, \mathrm{I\mbox{-}295}, \mathrm{I\mbox{-}395},
\mathrm{I\mbox{-}495}, \mathrm{I\mbox{-}66}, \mathrm{I\mbox{-}95}\}$.

We adopt a combination of traffic speed readings in \textit{detection time}
and \textit{early verification time} $\mathdata{F}=\{\v_{s}(\tau_{v}-h), ...,
\v_{s}(\tau_{v}+t)\}$ as the training features. For each traffic incident
subcollection $\mathdata{I}_{r}^{+}$, we construct the training input $\X_{r}$
and the label $\Y_{r}$. The problem is then formulated as solving the mapping:

\begin{equation}
F_r(\X_{r}) \rightarrow \Y_{r}
\label{equa:mapping}
\end{equation}
where $\X_{r}\in \mathbb{R}^{n_r\times p}, p=h+t; \Y_{r}\in \mathbb{R}^{n_r}$.
$n_r$ is the number of traffic incident records for one arterial road; $p$
represents the feature dimension of the training data, which is a combination
of the detection time and the verification time; $F_r$ is the learning model
for inferring the traffic incident duration in the subcollection
$\mathdata{I}_{r}^{+}$.

Consider that our problem is to predict the duration of the traffic
incidents if there is a historical traffic speed reading for the
corresponding collection of target traffic incidents $\mathdata{I}^{+}$, then
it fits into the scope of the regression problem. For instance, learning the
function $F_{r}$ can be modeled as a regression problem with a least square
loss function, and the model parameters $\w_r$ can be learned by solving the
following optimization problem:
\begin{equation}
 	\argmin{\W_r}\mathcal{L}_{r} = \ls{\X_{r}\W_{r} - \Y_{r}}/n_r + \lambda_{\W}\lnorm{\W_r}{1}
 	\label{equa:regression}
\end{equation}
where $\lambda_{\W}$ controls the sparsity of the grouped features, $n_r$ is
the total number of data points in $\mathdata{I}^{+}$. Moreover, as inspired
by the spatial correlations of traffic incidents introduced by the
connectivity between road segments, the subproblem $F_{r}$ defined in
Section~\ref{sec:probstate} to a regression problem under a multi- task
learning framework. The proposed model should be encouraged  to capture hidden
patterns among road segments and to maintain sparsity in feature space.
Mathematically, this consideration inspires us to use the
$\ell_{2,1}$ norm~\cite{argyriou2008convex} to perform joint feature
selection:
\begin{equation}
	\argmin{\W}\sum_{r=1}^{|\Phi|} \ls{\X_{r}\W_r - \Y_{r}}/n_r + \lambda_{\W}\lnorm{\W_r}{2,1}
	\label{equa:group_assign}
\end{equation}
where each column of $\W$, which represented by $\W_r$, denotes the model
parameters for $F_r$. In this way, we can further model the relatedness among
the road segments with parameter matrix $\W$. The overview of the
\textit{TITAN} model is represented in Figure~\ref{fig:mod_view}. The
following subsections address the details of the constraints on orthogonality
and spatial connectivity. 

%% file: sections/Models.tex
\section{Model} 
\label{sec:models}

To identify the critical temporal features for traffic incident duration
prediction, orthogonal constraints are applied to the \textit{TITAN} model; to
properly model the correlations between the traffic incidents based on the
connectivity between the arterial roads, we apply a multi-task learning
framework while designing the model. 

\subsection{Group Feature Learning}
\label{sec:sec:group_const}

In the studies of Traffic Incident Management (TIM), one important task is to
identify the key response time points and periods of traffic incidents. Assume
that a two-vehicle collision occurs at 5:15 pm on the road segment of
\textit{Interstate 66}, based on the traffic speed readings from the traffic
sensor, the transportation agencies want to learn how much impact the traffic
incident will introduce to the local transportation system in terms of
duration in time. The traffic speed readings of 5 minutes and 15 minutes after
the traffic incident’s occurrence play an important role in predicting the
duration of the traffic incident.

\textbf{Definition II}\label{def:def3}: 
\textit{Groups of key time points for a traffic incident}. 
The group assignment information is represented in a vector, and the $i$th
group of time points is denoted by $\q_i\in\mathbb{R}^p$. If the $j$th time
point feature belongs to this group, then the $j$th component of $\q_i$ is
non-zero and the relative magnitude represents the `importance' of the feature
in this group. For training data $\X_r$ for one specific road segment, the new
features generated by the group assignment is given by $\X_r\q_i$. Assume that
there are $k$ groups of features and the group structure is denoted by
$\Q=[\q_1, \q_2, ..., \q_k]$, and the generalized new features are given by
$\X_r\Q$. To assign physical meaning to each generated group, the elements
of $\Q$ have to be non-negative. 



The new model vector for the grouped features is denoted by
$\w_r\in\mathbb{R}^k$. The resulting formulation of the key feature group
identification problem is then defined by:
\begin{equation}
    \begin{aligned}
        \argmin{\Q, \W} \mathcal{L} = & \sum_{r=1}^{|\Phi|}\ls{\X_{r}\Q\W_{r} - \Y_{r}}/n_r + \lambda_{\W}\lnorm{\W}{2,1}\\
        \mathrm{s.t.}\text{\ } & \Q\geq0,\lone{\q_i}\leq\theta, i=1, ..., k,
    \end{aligned}
    \label{equa:spacity}
\end{equation}
where $\theta$ the parameter that controls the sparsity of each assigned group
in $\Q$. The $\ell_1$-norm in the constraint determines the length of the
column in $\Q$ to be $\theta$, which makes the group matrix $\Q$ easy to be
interpreted. 

By solving Equation~\ref{equa:spacity}, the model learns the group structure
of the data features. However, the features may be largely overlapped because
the proposed constraint does not consider any restrictions on feature
overlapping. Such group overlapping is not ideal in our problem setting of
traffic incident duration prediction problem. Because our selection of
features is based on a time sequence of traffic speed readings, the
consecutiveness of the features always provides a physical meaning.

In the research of traffic incident management, the lifetime of an incident
generally consists of five different stages: incident detection, verification,
response, clearance, and recovery. Because all stages do not overlap with each
other, we impose the orthogonal constraints $\q_i^T\q_j=0$ to control the
overlapping conditions among the groups.  The original non\-negative
constraint $\Q\geq 0$ between all $i$,$j$ is also applied. In terms of
simplicity and interpretation, we normalize the group assignments and assume
that the columns of $\Q$ are of length 1 for $\ell_2$ norm.  The constraint
can further be expressed by $\Q^T\Q=\I$. We use the $\ell_1$ norm
regularization to control the sparsity on $\Q$. The improved formulation of
group feature learning can be given by:

\begin{equation}
    \begin{aligned}
        \argmin{\Q, \W} \mathcal{L} = & \sum_{r=1}^{|\Phi|}\ls{\X_{r}\Q\W_{r} - \Y_{r}}/n_r\\
        & + \lambda_{\W}\lnorm{\W}{2,1} + \lambda_{\Q}\lone{\Q}\\
        \mathrm{s.t.}\text{\ } & \Q^T\Q=\I, \Q\geq0
    \end{aligned}
    \label{equa:group_const}
\end{equation}

\subsection{Spatial Connectivity in Feature Space}
\label{sec:sec:multi_const}

In real-world transportation systems, different road segments are spatially
related by intersections or interchanges. That is, two or more road segment
may share similar traffic speed pattern during the traffic incidents. For
instance, traffic congestion on \textit{Interstate 495} could not only cause
traffic pattern change at local road segments but also lead to traffic
pattern change on other arterial roads that have close spatial correlations
(e. g. \textit{Interstate 66} and \textit{US Route 7}). This spatial
relatedness caused by network failure
cascade~\cite{su2014robustness,kwee2018traffic} results in similar traffic
speed fluctuations; therefore, a similar pattern of traffic incident
durations. 

\begin{figure}[htpb!]
    \centering
    \includegraphics[width=\linewidth]{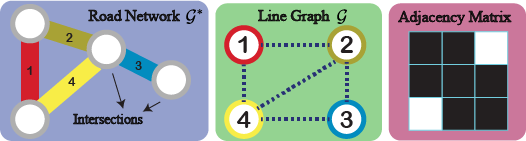}
    \caption{Road Segments Connectivity Shown by Adjacency Matrix. The left figure shows an example of the road network, the edges represent the road and the vertices represent the intersections; the middle figure shows the converted line graph of the road network, the vertices represent the roads; the right figure shows the adjacency matrix generated from the line graph. }\label{fig:g_b}
\end{figure}

\begin{figure*}[htpb!] 
\centering
\includegraphics[width=\linewidth]{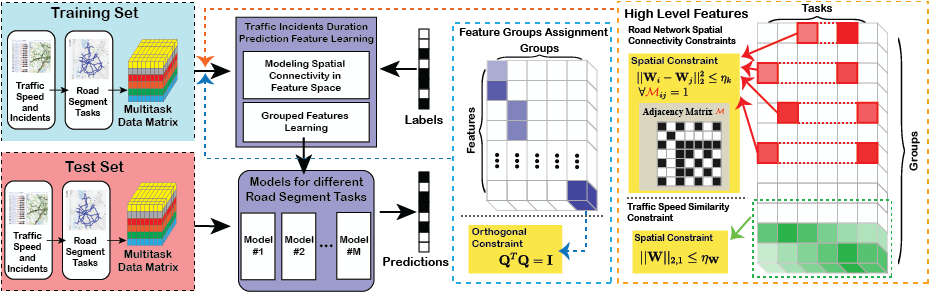} 
\caption{ A Schematic
View of the Traffic Incident Duration Prediction Model (TITAN). Similarities 
among temporal features are modeled by two major factors: spatial
connectivity between arterial roads and the orthogonal constraint on $\Q$. In
particular, arterial roads connectivity constraints encourage the model to
decrease differences between spatially related arterial roads in feature space.
The orthogonal constraint encourages the model to identify groups of critical
temporal features that are most influential to the prediction results. }
\label{fig:mod_view} 
\end{figure*}

\textbf{Definition III}\label{def:def2}: 
\textit{Traffic incident spatial correlations}.
With prior knowledge such as the road network connectivity, we assume that the
traffic incidents are spatially correlated with each other. Given a road
network $\mathdata{G}^{*}=(\mathdata{V}^{*}, \mathdata{E}^{*})$, where the
vertices set $\mathdata{V}^{*}$ represents the union collection of the
intersections and interchanges, and the edges set $\mathdata{E}^{*}$
represents the collection of roadblocks. In order to model the connectivity
of the road network, we transform the original road network graph
$\mathdata{G}^{*}$ to its line graph $\mathdata{G}=\mathsf{L}
(\mathdata{G}^{*})=(\mathdata{V}, \mathdata{E})$, where the vertices set
$\mathdata{V}$ represents the roads, and the edges set $\mathdata{E}$
represents the connectivity between roads. The adjacency matrix $\mathdata{M}$
of the line graph $\mathdata{G}$ reflects the overall connectivity of the
roads. The roads connectivity and the line graph transformation is shown in
Figure~\ref{fig:g_b}.  Mathematically, we improve the model with
constraints on parameters among different tasks:
\begin{equation}
    \begin{aligned}
    	\argmin{\Q, \W} \mathcal{L} = & \sum_{r=1}^{|\Phi|} \ls{\X_{r}\Q\W_r - \Y_{r}}/n_r + \lambda_{\W}\lnorm{\W}{2,1}\\
        \mathrm{s.t.}\text{\ } & \ls{\W_{i} - \W_{j}}\leq \eta_k, \eta_k\geq0, \forall \mathdata{M}_{ij}=1
    \end{aligned}
    \label{equa:multi_const}
\end{equation}
where each constraint with $\eta_k$ forces the Euclidean distance between
model parameters for a specific pair of road segments to be within a range. As
defined in Section~\ref{sec:probstate}, $\mathdata{M}$ is the adjacency matrix
that models the connectivity between road segments.

Combining the models represented by Equations~\ref{equa:group_const}
and~\ref{equa:multi_const}, we obtain our proposed \textit{TITAN} model. By
moving the non-trivial constraints that are correlated to spatial connectivity
into the objective function, we can obtain an equivalent regularized problem,
which is easier to solve:

\begin{equation}
    \begin{aligned}
    	\argmin{\Q, \W} & \sum_{r=1}^{|\Phi|}\ls{\X_{r}\Q\W_{r} - \Y_{r}}/n_r + \lambda_{\W}\lnorm{\W}{2,1} \\
        & + \lambda_{\Q}\lone{\Q} + \frac{1}{2}\sum_{ij}\mathdata{M}_{ij}\cdot\lambda_k\ls{\W_{i} - \W_{j}}\\
        \mathrm{s.t.}\text{\ } & \Q^T\Q=\I, \Q\geq0
    \end{aligned}
    \label{equa:tidm2}
\end{equation}
where $\lambda_k$ is trade-off penalty balancing the value of the loss
function and the regularizers. $\mathdata{M}$ is the adjacency matrix
representing the road connectivity; $\mathdata{M}_{ij}\in \{0,1\}$ denotes
the connectivity information between the $i$-th road and the $j$-th road.
Because the line graph $\mathdata{G}$ for road segments is undirected, the
corresponding adjacency matrix $\mathdata{M}$ is a symmetric matrix. The
coefficient $\frac{1}{2}$ is introduced to eliminate the repeatedly added
lower triangular matrix. 

%% file: sections/Algorithm.tex
\section{Parameter Learning for \textit{TITAN}}%
\label{sec:algorithm}

The objective function in Equation~\ref{equa:tidm2} is multi-convex and the
regularizer $\ell_{2,1}$ is non-smooth. This increases the difficulty of
solving this problem. A traditional way to solve this kind of problem is to
use proximal gradient descent. But this approach is slow to converge.
Recently, the alternating direction method of multipliers
(ADMM)~\cite{boyd2011distributed} has become popular as an efficient algorithm
framework which decouples the original problem into smaller and easier to
handle subproblems. Here we propose an ADMM-based Algorithm~\ref{alg:admm}
which can optimize the proposed models efficiently. In particular,
primal variables are updated on Line 4, dual variables on Line 5, and Lagrange
multipliers on Line 6. Line 7 calculates both primal and dual residuals.

\begin{algorithm}[htpb]
    \KwIn{$\X, \Y$}
    \KwOut{$\W, \Q$}
    Initialize $\W^{(0)}$, $\Q^{(0)}$, $\U_{\W}^{(0)}$, $\U_{\Q}^{(0)}$, $\mathbf{\Lambda}_{1}^{(0)}$, $\mathbf{\Lambda}_{2}^{(0)}$, $\mathbf{\Lambda}_{3}^{(0)}$\;
    Initialize $\rho = 1$, $\epsilon^{p} > 0, \epsilon^{d} > 0$, $\mathrm{MAX\_ITER}$\;
    \For{$k = 1 : \mathrm{MAX\_ITER}$}
    {%
        Update $\W^{(k)}$, $\Q^{(k)}$ with BCD using Equations~\ref{equa:grad_s} and ~\ref{equa:grad_G}\;  
        Update $\U_{\W}^{(k)}$ and $\U_{\Q}^{(k)}$ with Equations~\ref{equa:update_dual}\;
        Update $\mathbf{\Lambda}_{1}^{(k)}$, $\mathbf{\Lambda}_{2}^{(k)}$, and $\mathbf{\Lambda}_{3}^{(k)}$ with Equations~\ref{equa:update_lm}\;
        Compute $p$ and $d$ by Equations~\ref{equa:update_residuals}\;
        \If{$p < \epsilon^{p}$ and $d < \epsilon^{d}$}
        {%
            break\;
        }
    }
    \caption{An ADMM-based solver for TITAN.}\label{alg:admm}
\end{algorithm}


\subsection{Augmented Lagrangian Scheme}%
\label{sub:augmented_lagrangian_scheme}

First, we introduce an auxiliary variable $\U_{\Q} = \Q$ and $\U_{\W} = \W$
into the original problem~\ref{equa:tidm2} and obtain the following equivalent
problem:
\begin{equation}\label{equa:auxiliary_variables}
    \begin{aligned}
        \argmin{\Theta} & \sum_{r=1}^{|\Phi|}\ls{\X_{r}\Q\W_{r} - \Y_{r}}/n_r + \lambda_{\W}\lnorm{\U_{\W}}{2,1} \\
        & + \lambda_{\Q}\lone{\U_{\Q}} + \frac{1}{2}\sum_{ij}\mathdata{M}_{ij}\cdot\lambda_k\ls{\W_{i} - \W_{j}}\\
        \mathrm{s.t.}\text{\ } & \U_{\Q} = \Q, \U_{\W} = \W, \Q^T\Q=\I, \Q\geq0
    \end{aligned}
\end{equation}
where $\Theta = \{\W, \Q, \U_{\W}, \U_{\Q}\}$ is the set of variables to be
optimized. Then we transform the above problem into its augmented Lagrangian
form as follows:
\begin{equation}\label{equa:lagrange}
    \begin{aligned}
        & \argmin{\Theta} \sum_{r=1}^{|\Phi|}\ls{\X_{r}\Q\W_{r} - \Y_{r}}/n_r + \lambda_{\W}\lnorm{\U_{\W}}{2,1}\\
        & + \lambda_{\Q}\lone{\U_{\Q}} + \sum_{ij}\mathdata{M}_{ij}\cdot\lambda_k\ls{\W_{i} - \W_{j}}\\
        & + \langle\mathbf{\Lambda}_1, \W-\U_{\W}\rangle + \langle\mathbf{\Lambda}_2, \Q-\U_{\Q}\rangle + \langle\mathbf{\Lambda}_3, \I-\Q^T\Q\rangle\\
        & + \frac{\rho}{2}\lfrobsq{\W-\U_{\W}} + \frac{\rho}{2}\lfrobsq{\Q-\U_{\Q}} + \frac{\rho}{2}\lfrobsq{\I-\Q^T\Q}
    \end{aligned} 
\end{equation}
where $\mathbf{\Lambda}_1$, $\mathbf{\Lambda}_2$, and $\mathbf{\Lambda}_3$ are
the Lagrangian multipliers. With this step, we decouple the original problem
into two easier to handle problems in which seven variables $\W$,
$\Q$, $\U_{\W}$, $\U_{\Q}$, $\mathbf{\Lambda}_1$, $\mathbf{\Lambda}_2$, and
$\mathbf{\Lambda}_3$ will be optimized individually. Note that the coefficient
$\frac{1}{2}$ is omitted according to the optimization problem, and $\lfrob{\cdot}$ is the Frobenius norm. 

\subsection{Parameter Optimization}
\label{sub:parameters_optimization}

The Lagrangian form in Equation~\ref{equa:lagrange} is separated based on the primal variables and the dual variables, where the problem of solving the primal variables $\W$ and $\Q$ is smooth and convex:

\begin{equation}\label{equa:primal_var}
    \begin{aligned}
        & \argmin{\W, \Q} \sum_{r=1}^{|\Phi|}\ls{\X_{r}\Q\W_{r} - \Y_{r}} + \sum_{ij}\mathdata{M}_{ij}\cdot\lambda_k\ls{\W_{i} - \W_{j}}\\
        & + \langle\mathbf{\Lambda}_1, \W-\U_{\W}\rangle + \langle\mathbf{\Lambda}_2, \Q-\U_{\Q}\rangle + \langle\mathbf{\Lambda}_3, \I-\Q^T\Q\rangle\\
        & + \frac{\rho}{2}\lfrobsq{\W-\U_{\W}} + \frac{\rho}{2}\lfrobsq{\Q-\U_{\Q}} + \frac{\rho}{2}\lfrobsq{\I-\Q^T\Q}
    \end{aligned} 
\end{equation}

\subsubsection{Update $\W$}
\label{subsub:update_W}
We define Equation~\ref{equa:primal_var} as objective function $\mathcal{Q}$
which is multi-convex. In particular, $\mathcal{Q}$ of $\W_{r}$ is convex
where all other $\W_{r' \neq r}$ are fixed. This kind of problem can be
decoupled into subproblems using block coordinate descent
(BCD)~\cite{xu2013block}, in which each $\W_{r}$ is updated by solving the
following sub-optimization problems:

\begin{equation}
    \W_{r} \leftarrow \argmin{\W_{r}} \mathcal{Q}. 
\end{equation}
$\mathcal{Q}$ is smooth and convex for each $\W_{r}$ and can be solved by
gradient descent as follows: 
\begin{equation}\label{equa:grad_s}
    \frac{\partial \mathcal{Q}}{\partial \W_{i}} =
        \mathcal{P}(i) + 2\sum_{ij}\mathdata{M}_{ij}\cdot\lambda_k(\W_{i} - \W_{j})
\end{equation}
where according to the BCD algorithm, the $\partial\mathcal{Q}_{\W}/\partial
\W_{i}$ is calculated in sequence, from $i=1$ to $k$. And the
$\mathcal{P}(r)$ is defined as follows:
\[
    \begin{aligned}
        \mathcal{P}(r) = 2\Q^T\X_r^T(\X_{r}\Q\W_{r} - \Y_{r}) + \mathbf{\Lambda}_{1}^{r} + \rho(\W_{r} - \U_{\W}^{r})
    \end{aligned}
\]
where $\mathbf{\Lambda}_{1}^{r}$ and $\U_{\W}^{r}$ are the $r$-th columns of
the corresponding Lagrangian multiplier and dual variable.

\subsubsection{Update $\Q$}
\label{subsub:update_Q}

Similarly, the objective function $\mathcal{Q}$ of $\Q$ is also smooth and
convex. Because there are no constraints defined between the columns of $\Q$,
the problem can be solved by gradient decent directly based on the objective
function~\ref{equa:primal_var}, and the gradient of $\Q$ is calculated by:
\begin{equation}\label{equa:grad_G}
    \begin{aligned}
        \frac{\partial \mathcal{Q}}{\partial \Q} = & 2\sum_{r=1}^{|\Phi|}\X_{r}^T(\X_{r}\Q\W_{r} - \Y_{r})\W_{r}^T + \mathbf{\Lambda}_{2}\\
        & + 2\Q\mathbf{\Lambda}_{3} + \rho(\Q - \U_{\Q}) + 2\rho\Q(\Q^T\Q - \I)
    \end{aligned}
\end{equation}
and the primal variable $\Q$ is then updated with a step size of $\alpha$:
\begin{equation}\label{equa:update_G}
    \Q^{+} \leftarrow \Q - \alpha\cdot\frac{\partial \mathcal{Q}}{\partial \Q}
\end{equation}
Now that the primal variable $\W$ is taken care of, the dual variable $\U_{w}$ is
updated as follows:
\begin{equation}\label{equ:update_dule}
   \U_{w}^{+} \leftarrow \argmin{\U_{w}} \lambda_{5}\lnorm{\U_{w}}{2, 1}
   + \frac{\rho}{2}\ls{\U_{1} + \W - \U_{w}}.
\end{equation}
Note that this problem is the definition of proximal
$\prox_{f_{1}, 1/\rho}(\U_{1} + \W)$, where $f_{1}$ is the non-smooth
function $\lambda_{5}\lnorm{\U_{w}}{2, 1}$. The proximal operator can be solved
efficiently using~\cite{parikh2014proximal}.

\subsubsection{Update Dual Variables}
\label{subsub:update_dual}

Now that primal variables $\Q$ and $\W$ is taken care of, the dual variables
$\U_{\Q}$ and $\U_{\W}$ are updated as follows: 

\begin{equation}\label{equa:update_dual}
    \begin{aligned}
        \U_{\W}^{+} & \leftarrow \prox_{f_{1}, 1/\rho}(\mathbf{\Lambda}_{1} + \W),\\
        \U_{\Q}^{+} & \leftarrow \prox_{f_{2}, 1/\rho}(\mathbf{\Lambda}_{2} + \Q)
    \end{aligned}
\end{equation}
where $f_{1}$ is the non-smooth function $\lambda_{\W}\lnorm{\U_{\W}}{2, 1}$
and $f_2$ is the non-smooth function $\lambda_{\Q}\lnorm{\U_{\Q}}{1}$. The
proximal operator can be solved efficiently using proximal
operators~\cite{parikh2014proximal}.

Next, the Lagrangian multipliers $\mathbf{\Lambda}_{1}$,
$\mathbf{\Lambda}_{2}$, and $\mathbf{\Lambda}_{3}$ are updated as follows:
\begin{equation}\label{equa:update_lm}
    \begin{aligned}
        \mathbf{\Lambda}_{1}^{+} & \leftarrow \mathbf{\Lambda}_{1} + \rho(\W^{+} - \U_{\W}^{+})\\
        \mathbf{\Lambda}_{2}^{+} & \leftarrow \mathbf{\Lambda}_{2} + \rho(\Q^{+} - \U_{\Q}^{+})\\
        \mathbf{\Lambda}_{3}^{+} & \leftarrow \mathbf{\Lambda}_{3} + \rho(\Q^{+T}\Q^{+} - \I)\\
    \end{aligned}
\end{equation}
Finally, primal and dual residuals are calculated with: 
\begin{equation}\label{equa:update_residuals}
    \begin{aligned}
        p & = \ltwo{\W^{+} - \U_{\W}^{+}} + \ltwo{\Q^{+} - \U_{\Q}^{+}} + \ltwo{\Q^{+T}\Q^{+} - \I}\\
        d & = \rho\left(\ltwo{\U_{\W}^{+} - \U_{\W}} + \ltwo{\U_{\Q}^{+} - \U_{\Q}}\right).
    \end{aligned}
\end{equation}
where $p$ is primal residual, and $d$ is dual residual. 

%% file: sections/Experiment.tex
\section{Experiment}\label{sec:experiment}

In this section, we present the experiment environment, dataset introduction,
evaluation metrics and comparison methods, extensive experimental analysis on
predictive results, and discussions on the learner features.


\begin{table*}[htpb!]
	\large
    \centering
    \caption{Traffic Incident Duration Prediction Comparisons (RMSE (Min),
    MAE (Min), MAPE (\%))}\label{tab:eval}
\begin{tabular}{@{}lccc|ccc|ccc@{}}
\toprule
Method   & \multicolumn{3}{c}{I-270}                     & \multicolumn{3}{c}{I-295}                     & \multicolumn{3}{c}{I-395}                     \\ \cmidrule(l){2-10} 
         & RMSE          & MAE           & MAPE          & RMSE          & MAE           & MAPE          & RMSE          & MAE           & MAPE          \\ \midrule
Ridge       & 92.4709          & 76.4666          & 96.3826          & 89.1404          & 69.1273          & 87.3530          & 84.6881          & 65.5869          & 83.3106          \\
LASSO       & 90.8535          & 73.8732          & 90.3336          & 76.4372          & 58.8515          & 70.1599          & 72.4028          & 55.8695          & 68.8993          \\
SVR      & 87.8016          & 72.9036          & 88.7639          & 72.4579          & 53.9583          & 68.6843          & 68.4456          & 50.0854          & 62.6849          \\
nMTL     & \textbf{70.7942} & 59.9754          & 82.8141          & 55.4657          & 42.6052          & 55.3893          & 57.2953          & 43.3107          & \textbf{41.2034} \\
FeaFiner & 77.0080          & \textbf{57.5550}          & 81.4397          & 63.3036          & 50.1060          & 62.6381          & 51.6727          & 40.8695          & 47.4805          \\
TITAN    & 73.1291          & 59.5265 & \textbf{81.3789} & \textbf{46.0873} & \textbf{34.3043} & \textbf{52.9296} & \textbf{46.2329} & \textbf{38.9277} & 42.3794          \\ \hhline{==========}
Method   & \multicolumn{3}{c}{I-495}                     & \multicolumn{3}{c}{I-66}                      & \multicolumn{3}{c}{I-95}                      \\ \cmidrule(l){2-10} 
         & RMSE          & MAE           & MAPE          & RMSE          & MAE           & MAPE          & RMSE          & MAE           & MAPE          \\ \midrule
Ridge       & 69.9718          & 52.2384          & 81.2393          & 80.4118          & 62.5392          & 85.3443          & 76.0088          & 64.6172          & 80.1281          \\
LASSO       & 60.0119          & 48.5583          & 75.6027          & 68.0900          & 60.7429          & 77.9394          & 84.5617          & 58.7706          & 69.6493          \\
SVR      & 58.9676          & 46.7641          & 71.5021          & 72.7470          & 59.0808          & 71.1609          & 62.8689          & 54.7717          & 68.8999          \\
nMTL     & 52.5722          & 40.5422          & 63.6820          & 60.6244          & 48.4900          & 58.4887          & 57.1166          & 45.1327          & \textbf{49.4991} \\
FeaFiner & 56.3049          & 44.0023          & 44.9048          & 62.5098          & 50.4090          & 56.4438          & 55.6806          & 46.0073          & 56.0013          \\
TITAN    & \textbf{47.7131} & \textbf{31.7725} & \textbf{37.1649} & \textbf{53.7001} & \textbf{44.3786} & \textbf{40.9370} & \textbf{52.6403} & \textbf{40.5345} & 49.9848         
\\ \midrule
\end{tabular}
\end{table*}

\subsection{Experiment Setup}%
\label{ssub:experiment_setup}

\subsubsection{Experiment Environment}%
\label{subsub:experiment_environment}

We conducted our experiments on a machine with Intel Core i7-4790 3.6 GHz, the
computational power of this CPU is 4.13 Gflops per core. For real-world
traffic incident analysis problems, time requirements should be an important
factor. The most time-consuming process of our proposed \textit{TITAN} model
is at the training stage. The training stage learns the parameters for
temporal features $\W$ and the orthogonal groups of the temporal features
$\Q$. A matrix multiplication $\X\Q\W$ will generate the prediction rapidly.
In the validation and testing stages, our prediction for a single data point
is generated in less than $0.003$ seconds.

\subsubsection{Dataset and Feature Settings}%
\label{subsub:dataset_and_feature}

We evaluate our proposed Traffic Incident Duration Prediction model using two
real-world traffic data sources. \textbf{1) Traffic incident records with reported
duration}. We collect 43,923 records of traffic incidents in the year 2018
from three major transportation agencies in the Washington DC Metropolitan
area: Washington DC, Virginia State and Maryland State departments of
transportation. From the collected traffic incident records, we select 29,075
traffic incidents that take places on the six major arterial roads in the
region: $\mathrm{I\mbox{-}270}$, $\mathrm{I\mbox{-}295}$,
$\mathrm{I\mbox{-}395}$, $\mathrm{I\mbox{-}495}$, $\mathrm{I\mbox{-}66}$, and
$\mathrm{I\mbox{-}95}$. In the selected data frame, the time duration of the
traffic incidents are recorded in minutes, and we utilize the duration as the
ground truth. From the selected incidents 80\% of the records serve as the
training set, while the rest serve as the testing set. \textbf{2) INRIX
traffic speed data}. We leverage the traffic speed readings from the traffic
sensors as the training features. Given the location and verification time of
the traffic incidents, we collect traffic speed readings of nearby traffic
sensors. 

The connectivity of the road network determines the number of tunable
parameters in our \textit{TITAN} model. According to the selected arterial
roads in our experiment, seven hyperparameters can be tuned. During the
experiment, we observe that the value of the loss function is significantly
larger than regularizers, which means a large penalty should be used to
balance the loss function and the regularizers. 

\subsection{Comparison Methods}%
\label{sub:comparison_methods}

To evaluate the performance of the traffic incident duration prediction, 5
comparison methods are considered in our experiment: $\ell_2$ regulized linear
regression (ridge regression), $\ell_1$ regulized linear regression (LASSO),
support vector regression (SVR), Na\"ive multi-task learning model (nMTL), and
feature refiner method (FeaFiner). 

$\bullet$ \textbf{$\ell_2$ Regulized Linear Regression (Ridge)}~\cite{peeta2000providing}. Ridge regression is an extension for
linear regression. It's a linear regression model regulized on $\ell_2$ norm.
The $\lambda$ parameter is a scalar that controls the model complexity; the
smaller $\lambda$ is, the more complex the model will be. In our
implementation, $\lambda$ is searched from $\{10, 100\}$. This model only
considers the temporal features on duration prediction. No multi-task for
arterial road connectivity and grouped temporal features are considered. 

$\bullet$ \textbf{$\ell_1$ Regulized Linear Regression (LASSO)}~\cite{ramakrishnan2014beating,tibshirani1996regression}. This is a
classic way to conduct cost-efficient regressions by enforcing the sparsity of
the selected features. It has been proved to be effective in the field of event detection~\cite{ramakrishnan2014beating}. It includes a parameter $\lambda$ that trades off the
regularization term; typically, the larger this parameter is, the fewer the
selected features will be. In our experiment, $\lambda$ is searched from $\{1,
10, 100\}$. The feature configurations applied by this model is the same as
the ridge regression model. 

$\bullet$ \textbf{Support Vector Regression
(SVR)}~\cite{tibshirani1996regression}. Support vector regression provides
solutions for both linear and non-linear problems. In our experiment
implementation, we utilize non-linear kernel functions (RBF kernel) to find
the optimal solution for incident duration prediction problem. The model
parameters are selected with $c=1$ and $\epsilon=0.1$. This model considers
similar temporal features with ridge regression and LASSO methods, no
multi-task features for connectivity is considered. 

$\bullet$ \textbf{Na\"ive Multi-task Learning Model
(nMTL)}~\cite{zhao2016hierarchical}. We implement the fundamental settings of
the naive multi-task learning model for event detection. This comparison
method is regularized with $\ell_21$ constraint between tasks. The training
tasks of this model are split by the arterial roads. The correlations between
tasks are intuitively constrained by $\ell_2$ norm, and within each task, the
importance of the features are constrained by $\ell_1$ norm. The penalty
parameter $\lambda$ is searched from $\{1, 10, 100\}$. 

$\bullet$ \textbf{FeaFiner}~\cite{zhou2013feafiner}. FeaFiner regression model
with a capability of learning feature clusters. This method learns an optimal
sparse feature grouping for general regression problems. However, there are no
multi-task properties supported. In our implementation of this method, we
apply this method on the complete set of traffic incidents, and the target
feature is selected to be the temporal features. In the parameter
initialization, we select the parameter $k=30$ for the k-Mean clustering. 

\subsection{Evaluation Metrics}%
\label{sub:evaluation_metrics}

To quantify and validate model performance on traffic incident duration
prediction, we adopt root mean squared error (RMSE), mean absolute error
(MAE), and mean absolute percentage error (MAPE). These metrics are widely
utilized in the field of traffic duration prediction
studies~\cite{li2018overview,khattak2016modeling,park2016interpretation,zou2016application},
it reflects the predictive performance of the proposed model. Equations~\ref{equa:rmse},~\ref{equa:mae}, and~\ref{equa:mape} represent the calculations of the selected evaluation metrics: 

\begin{equation}
RMSE(\boldsymbol{y}, \hat{\boldsymbol{y}})=\sqrt{\frac{1}{N} \sum_{i=1}^{N}\left(y_{i}-\widehat{y}_{i}\right)^{2}}
\label{equa:rmse}
\end{equation}

\begin{equation}
MAE(\boldsymbol{y}, \hat{\boldsymbol{y}})=\frac{1}{N} \sum_{i}^{N}\left|y_{i}-\widehat{y}_{i}\right|
\label{equa:mae}
\end{equation}

\begin{equation}
MAPE(\boldsymbol{y}, \hat{\boldsymbol{y}})=\frac{1}{N} \sum_{i}^{N}\left|\frac{y_{i}-\widehat{y}_{i}}{y_{i}}\right|
\label{equa:mape}
\end{equation}
where $N$ is the total number of records; $\boldsymbol{y}$ is the predicted
traffic incident durations represented in vector; $\hat{\boldsymbol{y}}$ is the
ground truth value of the corresponding record, which is also represented in
vector. $y_{i}$ and $\widehat{y}_{i}$ are the $i^{th}$ predicted result and
the $i^{th}$ ground truth value respectively. 

\subsection{Incident Duration Prediction Analysis}%
\label{sub:prediction_results}

\subsubsection{\textit{TITAN} Performance Analysis on Spatial Connectivity}
\label{subsub:titan_connect}
Table~\ref{tab:eval} summarizes the comparisons of our proposed method to the
competing methods for the task of traffic incident duration prediction. From
the experimental results, we can justify our application of a multi-task
learning framework for predicting the incident duration. In general,
\textit{TITAN} outperforms the single task models (LR, SVR, and FeaFiner) on
RMSR, MAE, and MAPE. This result shows that the spatial correlations between
the road segments can improve the performance of the traffic incident duration
prediction. The \textit{TITAN} model outperforms the nMTL on RMSE and MAE.
These results demonstrated that for the traffic incident duration prediction
problem, only $\ell_1$ regularizers is insufficient, detailed spatial
connectivity between the road segments should also be considered.

\subsubsection{\textit{TITAN} Performance Analysis on Feature Groups Learning}
\label{subsub:titan_feagroup}
\textit{TITAN} Performance Analysis on Feature Groups Learning. Among the 
comparison methods, the FeaFiner~\cite{zhou2013feafiner} method considers the
orthogonal constraint that is capable of grouping low-level features into a
high-level feature representation. The original FeaFiner applies the Ridge and
LASSO as the original problem settings. Thus, the results presented in
Table~\ref{tab:eval} can be categorized by whether the orthogonal constraints
are considered or not. The methods consider orthogonal constraints are
FeaFiner and \textit{TITAN}; the methods do not consider the orthogonal
constraint are Ridge, LASSO, and SVR. By comparing these two categories, we
learn that the overall performance of the methods consider the orthogonal
constraint is better than the methods do not consider the orthogonal
constraint. However, the overall performance increase is not as significant as
the performance increase from the spatial connectivity constraint introduced
by the framework of multi-task learning. 

\subsubsection{Performance Analysis between Training Tasks}
\label{subsub:task_perform}
The results in Table~\ref{tab:eval} show that the model performance for traffic
incident duration prediction is not the same across different road segments.
For instance, the prediction performances of all the comparison methods on
highway $\mathrm{I\mbox{-}270}$ only have slight differences between each
other. This is because the highway $\mathrm{I\mbox{-}270}$ only has one
spatial connectivity to the rest of the road segments, and the constraint of
Euclidean distance for $\mathrm{I\mbox{-}270}$ only shares a limited
connection between the other columns of the feature matrix $\W$. In contrast,
our model for the highway $\mathrm{I\mbox{-}495}$ outperforms the comparison
methods, because the subtask for $\mathrm{I\mbox{-}495}$ shares feature
similarity with all other subtasks. 

\begin{figure}[ht]
\centering
\subfigure[TITAN w/o Orth. Const.]{\label{fig:titan_wo}\includegraphics[width=41mm]{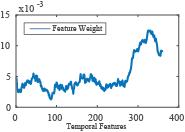}}
\subfigure[TITAN with Orth. Const.]{\label{fig:titan_wi}\includegraphics[width=41mm]{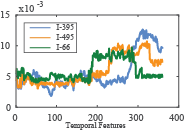}}
\caption{Feature Learning Results on $\Q$}
\label{fig:g_const}
\end{figure}

\begin{figure*}[htpb!] 
\centering
\subfigure{\includegraphics[width=57mm]{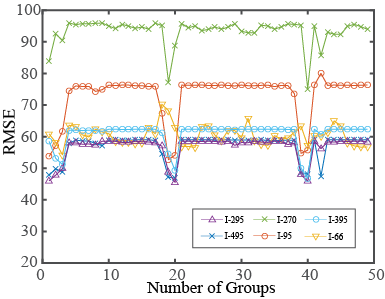}}
\subfigure{\includegraphics[width=57mm]{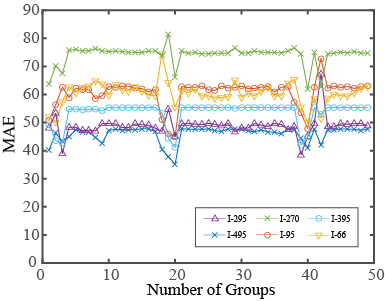}}
\subfigure{\includegraphics[width=57mm]{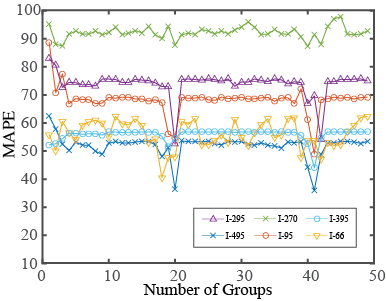}}
\caption{Illustration of how the number of grouped temporal features will affect the performance of the TITAN model. The performance is evaluated in terms of RMSE, MAE, and MAPE respectively. }
\label{fig:case_metrics}
\end{figure*}

\subsection{Feature Groups Assignment Analysis}%
\label{sub:feature_analysis}
The orthogonal constraint ensures the proposed model to learn a group of
highlighted features that play an important role in predicting the traffic
incident durations. In our experiment, we also study the results of the
learned group features empirically. In the experiment, we set the number of
groups to be 10, and we also apply two conditions: 1) \textit{TITAN} with
orthogonal constraint and 2) \textit{TITAN} without orthogonal constraint.
Figure~\ref{fig:g_const} shows the learned feature groups assignments for both
experimental conditions. We can find the learned features with orthogonal
constraint overlap less than the learned feature assignment without orthogonal
constraint. 

While experimenting without the orthogonal constraint, we found that the model
has a preference for grouping the low-level features into one feature
assignment for every group $\q_i$. Figure~\ref{fig:titan_wo} shows the single
feature group assignment for the model without orthogonal constraint. From
Figure~\ref{fig:titan_wo}, we can find that for the model without orthogonal
constraint, temporal features with higher indexes are assigned with higher
weights (>300). This result is reasonable because this can be interpreted as
the duration of the traffic incident can be better inferred with the most
recent traffic speed readings.

To compare with the model with orthogonal constraint,
Figure~\ref{fig:titan_wi} shows the learned feature group assignment for
several subtasks. We can find the most weighted feature group by checking the
weights in the learned variable $\W$. For example, in
Figure~\ref{fig:titan_wi}, we demonstrate top weighted group for three
subtasks ($\mathrm{I\mbox{-}495}$, $\mathrm{I\mbox{-}66}$, and
$\mathrm{I\mbox{-}395}$). From Figure~\ref{fig:titan_wi}, we find that the top
assigned feature group for different arterial roads differ from each other
slightly. This result shows that the most critical temporal features for
predicting the traffic incident duration for different roads differ. This
observation is valuable for the transportation operators and first responders.
In Figure~\ref{fig:titan_wi}, we can observe that the high-level features of
the subtask $\mathrm{I\mbox{-}495}$ have a shift comparing to the subtask of
$\mathrm{I\mbox{-}395}$. The 10 minutes’ shift indicates that to predict the
duration of an incident on $\mathrm{I\mbox{-}495}$, the traffic speed readings
of 10 minutes in advance have higher importance.

\subsection{Case Studies}%
\label{sub:case_studies}

During the experiments, several interesting facts revealed by using the
proposed approach were discovered. Here we discuss the details towards the
identification of the critical phases for traffic incidents and the influences
of the connectivity between the arterial roads.  

\subsubsection{Critical Phases Identification for Traffic Incidents}%
\label{subsub:critial_phase}

According to the experiment results on the correlations between the number of
groups and the performance of the \textit{TITAN} mode, we discover the optimal
number of groups for the temporal features. The physical meaning of the number
of groups in this experiment, corresponding to the number of phases will be
identified for the traffic incidents. As mentioned in Section I, the life
cycle of the traffic incident is conventionally grouped into five phases:
detection, verification, response, clearance, and recovery. Although such
grouping strategy is efficient in the perspective of transportation management
and operations, it cannot provide useful temporal feature grouping to predict
the traffic incident durations. From this experiment, we can study how the
performance of the \textit{TITAN} model will be affected with respect to the
number of feature groups. As shown in Figure~\ref{fig:case_metrics}, we
illustrate the RMSE, MAE, and MAPE obtained by varying the number of the
groups from 1 to 50; and the color-coded lines representing different arterial
roads in the experiment. From Figure~\ref{fig:case_metrics}, we learn that for
most of the arterial roads, the \textit{TITAN} model reaches the best
performance when the number of groups in the ranges of 18-20 and 40-43. This
experiment result indicates that the conventional five-phase definition of
traffic incident life cycle may not provide informative input to the traffic
incident duration prediction problems. 

\subsubsection{Influences of Arterial Road Connectivity}%
\label{subsub:influence_conn}

The performance differences between the arterial roads can be observed in
Figure~\ref{fig:case_metrics}. In Figure~\ref{fig:case_metrics}, the general
prediction performance of the arterial road Interstate 495 outperforms the
rest of the arterial roads, and the arterial road Interstate 270 has the worst
duration prediction results overall. This comparison result reveals that the
connectivity between different arterial roads plays an important role while
predicting the traffic incident duration. Because the more connection with
other arterial roads means the more information shared with other train tasks
in the training stage. The Interstate 495 intersections with all other
arterial roads, while the Interstate 270 only intersects with the Interstate
495. 

%% file: sections/Conclusion.tex
\section{Conclusion} 
This paper presents a novel traffic incident duration
prediction and feature learning model \textit{TITAN}. The proposed model is
designed based on the multi-task learning framework for prediction, and a
sparse feature learning framework for higher feature groups identification.
The proposed \textit{TITAN} model outperforms the existing traffic incident
duration prediction models because of two major advantages in model design: 1)
consideration of the connectivity between road segments within the urban road
networks; 2) the learned higher level features provide a better predictive
pattern for the problem of traffic incident duration prediction.  Extensive
experiments on real-world datasets with comparisons of the baseline methods
justify the performance of \textit{TITAN} model. By applying the orthogonal
constraint, the proposed model is capable of identifying groups of higher
level features which can be further considered as the critical evolution
stages of the traffic incident. Such functionality provided by our proposed
model is helpful for the transportation operators and first responders while
judging the influences of the traffic incidents. 